\documentclass[journal,twoside]{IEEEtran}

\usepackage[caption=false]{subfig}
\usepackage{cite}
\usepackage{url}
\usepackage{graphicx}
\usepackage[tbtags]{amsmath}
\usepackage{graphics}
\usepackage{epsfig}
\usepackage{amsthm, amssymb}
\usepackage{mathrsfs}
\usepackage{algorithm}
\usepackage{algorithmicx}
\usepackage{algpseudocode}
\usepackage[table,xcdraw]{xcolor}
\usepackage{array}
\usepackage{tabularx}
\usepackage{xspace}
\usepackage{bbm}
\usepackage{booktabs}
\usepackage{amsthm}

\title{Split Knowledge Distillation for Large Models in IoT: Architecture, Challenges, and Solutions}
\author{Zuguang~Li,~\IEEEmembership{Graduate Student Member,~IEEE},
Wen~Wu,~\IEEEmembership{Senior~Member,~IEEE},
Shaohua~Wu,~\IEEEmembership{Member,~IEEE},
Qiaohua~Lin,
Yaping~Sun,~and
Hui~Wang

\thanks{
Z. Li is with School of Electronics and Information Engineering, Harbin Institute of Technology, Shenzhen, 518055, China, and also with the Frontier Research Center, Peng Cheng Laboratory, Shenzhen, 518055, China (e-mail: lizuguang@stu.hit.edu.cn). 

W. Wu and Q. Lin are with the Frontier Research Center, Peng Cheng Laboratory, Shenzhen, 518055, China (email: \{wuw02, lingqh\}@pcl.ac.cn).

S. Wu is with the Guangdong Provincial Key Laboratory of Aerospace Communication and Networking Technology, Harbin Institute of Technology, Shenzhen, 518055, China (e-mail: hitwush@hit.edu.cn).

Y. Sun and H. Wang are with the Department of Broadband Communication, Peng Cheng Laboratory, Shenzhen, 518055, China (email: \{sunyp, wangh06\}@pcl.ac.cn).

This paper has been submitted to IEEE Internet of Things Magazine (under review).

}
}

\begin{document}

\maketitle
\begin{abstract}
   Large models (LMs) have immense potential in Internet of Things (IoT) systems, enabling applications such as intelligent voice assistants, predictive maintenance, and healthcare monitoring. However, training LMs on edge servers raises data privacy concerns, while deploying them directly on IoT devices is constrained by limited computational and memory resources. We analyze the key challenges of training LMs in IoT systems, including energy constraints, latency requirements, and device heterogeneity, and propose potential solutions such as dynamic resource management, adaptive model partitioning, and clustered collaborative training. Furthermore, we propose a split knowledge distillation framework to efficiently distill LMs into smaller, deployable versions for IoT devices while ensuring raw data remains local. This framework integrates knowledge distillation and split learning to minimize energy consumption and meet low model training delay requirements. A case study is presented to evaluate the feasibility and performance of the proposed framework.

\end{abstract}

\section{Introduction}

The emergence of large models (LMs), particularly large language models, has reshaped the landscape of artificial intelligence (AI), driving innovation across numerous domains. These revolutionary models, such as GPT, LLaMA, and their successors, have demonstrated exceptional capabilities in natural language understanding, human-like text generation, and decision-making~\cite{cui2024llmind}. Beyond their current success, LMs are increasingly recognized for their potential to empower the Internet of Things (IoT) system, where they can transform IoT devices into intelligent agents capable of understanding context, predicting user needs, and optimizing system performance. By leveraging their advanced inference and communication capabilities, LMs can facilitate seamless interactions and efficient operations within complex and interconnected IoT networks.

Despite their transformative capabilities, LMs are often too resource-intensive for deployment on IoT devices, which are constrained by limited computational power, memory, and energy resources~\cite{xiao2024efficient}. For example, practical industry LMs like GPT-3 175B require over 350 GB to store their parameters, which needs at least five NVIDIA H100 GPUs to accommodate this model~\cite{na2024understanding}. Therefore, the mainstream LMs are highly difficult to be deployed on IoT devices such as the Jetson platform, which have much more limited memory and computational resources.
To address this, knowledge distillation provides an effective solution by training a smaller student model using the outputs of a larger teacher model. 
This process transfers the teacher's knowledge while significantly reducing the model size and complexity. For example, student models such as MT-CoT, UniNER 7B, and ILD retain much of the teacher LM's performance while being lightweight enough to be deployed on IoT devices~\cite{yang2024survey}. By doing this, advanced AI functionalities can be achieved in resource-constrained environments, bridging the gap between LM capabilities and IoT limitations.

However, many private institutions, companies, or mobile carriers cannot use the dataset collected by users for distilling LMs due to data privacy and security concerns. To address this challenge, split learning has emerged as a promising solution that ensures data privacy while reducing the computational burden on IoT devices. In split learning, the initial layers of the model are executed locally on IoT devices, and only the intermediate activations stripped of raw data information are transmitted to the edge server, which processes the remaining layers of the model~\cite{wu2023split}. Although its advantages, leveraging split learning to distill LMs into small models for IoT applications still faces significant challenges. First, the energy consumption associated with transmitting intermediate activations and running computationally intensive layers on the edge server is substantial, particularly for resource-constrained IoT environments. Second, the efficiency of model training can be hindered by communication latency and the frequent synchronization required between devices and servers, which may slow down the overall training process and impact real-time applications. Addressing these issues is crucial to unlock the full potential of split learning for LMs in IoT systems.

In this article, we analyze the key challenges of deploying LMs in IoT systems, including energy constraints, latency requirements, and device heterogeneity, and propose potential solutions such as dynamic resource management, adaptive model partitioning, and clustered collaborative training. To address these challenges, we propose a split knowledge distillation framework that integrates the knowledge distillation and split learning to efficiently distill LMs into smaller, deployable models for IoT devices, while ensuring that raw data remains local to preserve privacy. This framework is designed to minimize energy consumption and meet low model training delay requirements. Finally, we validate the feasibility and performance of the proposed framework through a case study.

The rest of this article is organized as follows. We discuss the background in Section~\ref{sec: background}, and analyze the challenges and potential solutions in Section~\ref{sec: Challenges and Potential Solutions}. Section~\ref{sec: Proposed Framework} introduces the split knowledge distillation framework. A case study is presented in Section~\ref{sec: Case Study}.  Finally, Section~\ref{sec: Conclusion} concludes this paper. 

\section{Background} \label{sec: background}

\subsection{Large Model}

As a significant advancement in LMs, large language models are built on the transformer architecture, which uses a self-attention mechanism to efficiently capture long-range dependencies and understand world context~\cite{huang2023advancing}. This design allows LMs to process large datasets with parallel processing, setting them apart from earlier sequential models like recurrent neural networks (RNNs). This evolution has enabled breakthroughs in natural language processing tasks which require deep contextual understanding and scalability. 
Over time, driven by advances in hardware, software, and optimization techniques, LMs have evolved from models with millions of parameters to those with billions or even trillions of parameters, such as GPT-4 and LLaMA-3. 

\begin{figure*}[t]
	\renewcommand{\figurename}{Fig.}
	\centering
	\includegraphics[width=0.6\textwidth]{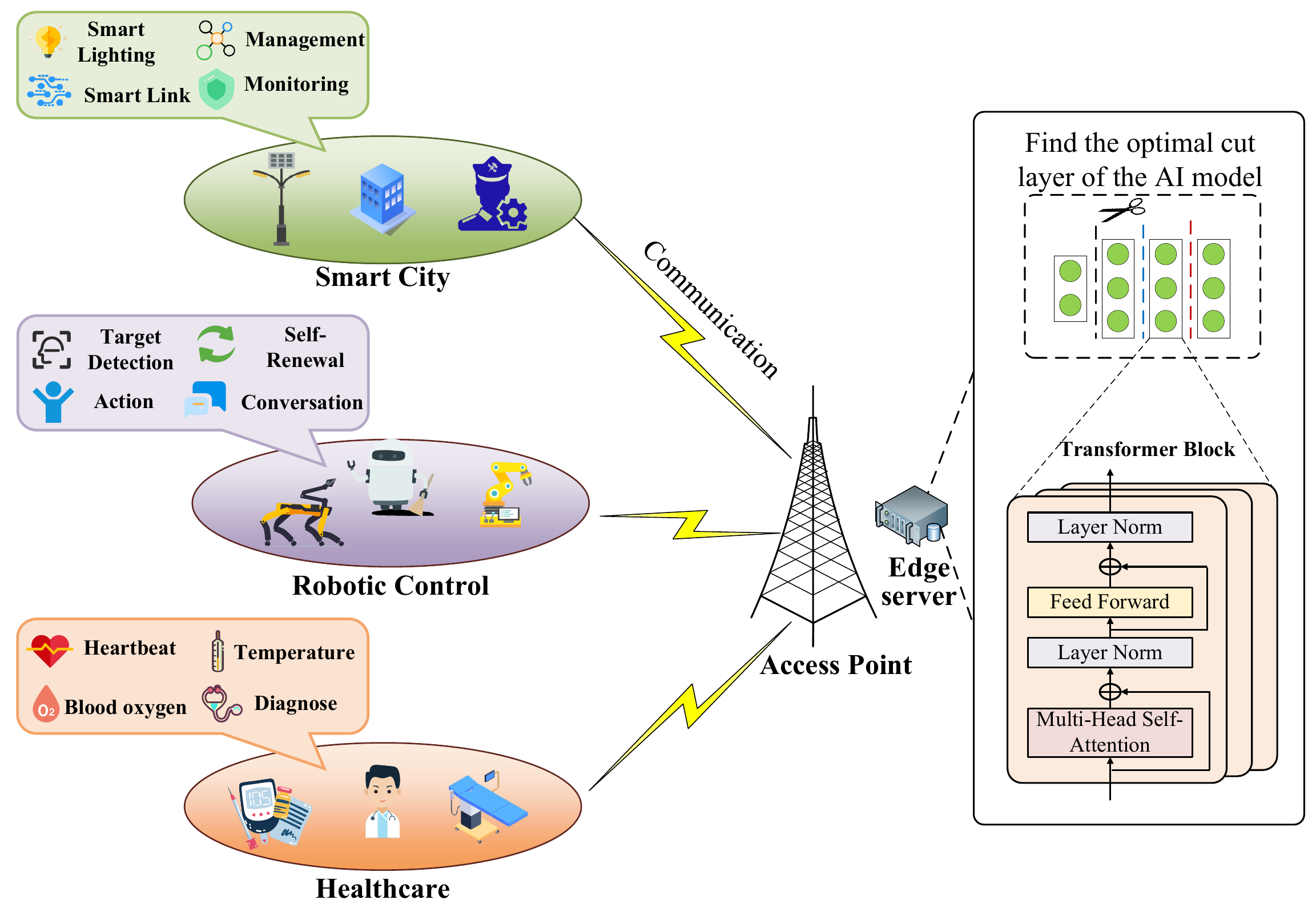}
	\caption{The application scenarios of LMs in the IoT scenario.}
	\label{fig: The application scenarios of LMs in IoTs}
    \vspace{-0.4cm}
\end{figure*}

Large amounts of data are generated from heterogeneous IoT devices, which is beneficial to users, businesses, and industries. However, much data could be unstructured and complex, presenting challenges for effective data processing. In this context, LMs have the potential to make IoT data more meaningful, mainly through their capabilities in natural language processing and understanding. 
When integrated into IoT systems, LMs can power applications in smart cities, robotic control, and healthcare, enabling intelligent decision-making, automation, and enhanced user interaction across diverse domains, as shown in Fig.~\ref{fig: The application scenarios of LMs in IoTs}. For example, in a Tactile IoT system used for remote robotic surgery, LMs can analyze sensor data, integrating medical expertise to assist in tasks like detecting tissue resistance during surgery, improving accuracy and safety~\cite{ahuja2021touchpose}. 
Conversely, LMs benefit from IoT by leveraging real-time, multi-modal, and rich data to enhance their understanding of advanced meaning and context-based decisions. The heterogeneous IoT devices provide continuously updated data, enabling LMs to adapt to changing environments and deliver more accurate and up-to-date responses.

For resource-constrained environments like IoT, deploying LMs presents significant challenges due to their size and computational demands. To address these issues, various optimization techniques are employed to reduce the model size and computational costs. Methods like quantization, which reduces the bit width of data, and pruning, which removes unnecessary weights, are commonly used to make LMs more efficient~\cite{xu2023survey}. Additionally, parameter-efficient fine-tuning methods, such as low-rank adaptation, focus on fine-tuning only critical parameters, enabling more efficient training with significantly reduced computational requirements~\cite{rao2024parameter}. Furthermore, knowledge distillation transfers the knowledge from an LM to a smaller one, which reduces the model's size, thereby making LMs more suitable for IoT devices.

\subsection{Knowledge Distillation}

To address the challenges of deploying LMs in constrained IoT environments, knowledge distillation is widely adopted. It plays a critical role in compressing advanced LMs like GPT-3 or LLaMA into compact versions without significant loss in capabilities. For example, as shown in Table~\ref{Table: representative knowledge distillation methods on LMs}, the GPT-3 model (700 GB) can be distilled into a student model like MT-CoT (12 GB) with a 58× compression rate while retaining 98\% of the teacher LM's performance. 
The distillation process begins by guiding the teacher LM to focus on a specific target domain or skill, achieved by employing structured prompts, precise instructions, or domain-specific datasets. This ensures that the teacher LM produces outputs demonstrating expertise in general cognitive skills (e.g., context following) or specialized fields (e.g., law and science).

During the training phase, the student model learns to approximate the behavior of the teacher LM by minimizing a loss function that combines two components: the Kullback-Leibler (KL) divergence between the teacher LM’s soft labels and the student model’s predictions, and the cross-entropy loss between the true labels and the student’s predictions. The soft labels provide richer information by encoding the teacher LM’s probability distribution over all possible outputs, helping the student generalize effectively. By optimizing this combined loss, the student model can mimic the teacher LM’s decision-making while operating with a smaller architecture. This enables the creation of lightweight models that maintain high accuracy and efficiency, suitable for IoT devices with limited computational resources and energy budgets.

\begin{table*}[]
	\scriptsize
    \renewcommand\arraystretch{1.5}
	\centering
	\caption{Comparison of Representative Knowledge Distillation Methods on LMs.}
	\label{Table: representative knowledge distillation methods on LMs}
        \centering
\begin{tabular}{ccccccc}
\hline
\textbf{Models} &
  \textbf{Model Size} &
  \textbf{Distillation Type} &
  \textbf{Student Model} &
  \textbf{Student Model Size} &
  \textbf{Compression Rate} &
  \textbf{Comparison with Teacher Model} \\ \hline
$\textit{GPT-3}_{text-davinci-002}$ & 700 GB & CoT & MT-CoT          & 12 GB  & 58$\times$ & 80.5/82.1 (98\% performance)  \\
$\textit{GPT-3.5}_{turbo-0301}$     & 400 GB & IF  & UniNER 7B       & 28 GB  & 14$\times$ & 41.7/34.9 (119\% performance) \\
GPT-3 175B             & 350 GB & CoT & Learn-to-Reason & 12 GB  & 29$\times$ & 62.1/76.1 (82\% performance)  \\
PaLM 60B               & 240 GB & CoT & PaD             & 3  GB  & 78$\times$ & 43.9/50.2 (88\% performance)  \\
LLaMA-3 13B            & 52 GB  & ICL & LLM-R           & 26 GB  & 2$\times$  & 68.8/64.6 (107\% performance) \\
Alpaca 7B              & 28  GB & IF  & LaMini-LM       & 3  GB  & 9$\times$  & 60.8/62.3 (98\% performance)  \\
BERT Large             & 1.36 GB & ICL & ILD             & 0.10 GB & 13$\times$ & 52.4/57.3 (91\% performance)  \\ \hline
\end{tabular}
\vspace{-0.4cm}
\end{table*}

\subsection{Split Learning}
Split learning has garnered significant attention as an effective solution for training AI models while preserving data privacy in resource-constrained environments, particularly in edge networks and IoT systems.  Existing works on split learning have demonstrated that the energy consumption in the model training can be effectively reduced through strategies such as cut layer selection, resource management, and architectural design.
Kang \textit{et al.} \cite{kang2017neurosurgeon} analyzed the per-layer execution time and energy consumption in different AI models, and then the optimal cut layer can be determined for the best latency or energy consumption.
Ayad \textit{et al.} \cite{ayad2023efficient} proposed a modified split learning approach to achieve the electrocardiography classification while reducing the communication overhead and computation workload.

Split learning has shown immense potential in various application domains, including autonomous driving, robotic control, and healthcare. In autonomous driving, split learning enables the efficient training and fine-tuning of AI models for tasks such as sensor fusion, object detection, and decision-making, all while minimizing energy consumption and ensuring real-time responsiveness~\cite{padaria2023traffic}. The framework facilitates adaptive model training for motion planning and environment interaction, addressing the energy constraints of edge devices commonly deployed in robotics. In healthcare, split learning has been applied to privacy-sensitive tasks such as medical image analysis and remote patient monitoring, where it reduces the computational burden on wearable or IoT devices while maintaining stringent data security and energy efficiency standards~\cite{ghosh2024split}. These applications highlight the versatility and transformative potential of split learning across energy-sensitive, latency-critical domains.

\section{Challenges and Potential Solutions} \label{sec: Challenges and Potential Solutions}

\subsection{Challenges}
\subsubsection{Limited Energy Consumption}
IoT devices often operate under strict energy constraints, which poses a significant challenge when deploying split knowledge distillation frameworks. Training and distilling LMs require frequent communication between IoT devices and edge servers, as well as substantial local and server-side computations. This energy-intensive process can quickly drain device batteries and increase operational costs, particularly in large-scale IoT networks. Moreover, high energy consumption limits the feasibility of real-time AI applications, such as robotic controlling or healthcare monitoring, where devices must operate continuously over extended periods without recharging.

The challenge is further compounded by the diverse energy capacities of IoT devices, leading to imbalanced workloads across the system. Devices with limited power may experience frequent interruptions or degraded performance, disrupting collaborative model training or inference. Addressing energy consumption is thus a critical requirement for achieving sustainable and reliable IoT deployments.

\subsubsection{Low Model Training Delay Requirement}
The second challenge in IoT applications is meeting the low model training delay requirement, as real-time processing is often crucial. Tasks such as smart city management, industrial automation, and healthcare monitoring demand quick decision-making, where delays in model training could significantly impact the system’s responsiveness and performance. Model training, particularly in the context of LMs, can introduce significant delays due to the complexity of AI models.

The training delay issue is exacerbated in split learning systems, where the training process is distributed between the edge server and IoT devices. This distribution introduces additional communication overhead and synchronization delays, further increasing the time required to complete each training round. Meeting low latency requirements while ensuring model accuracy is challenging, as optimizing the system for both speed and performance demands careful coordination of resources, network communication, and computation.

\subsubsection{Heterogeneous IoT Devices}
The third significant challenge in IoT systems is heterogeneous IoT devices that operate within these networks. These devices vary greatly in computational power, memory, and energy capabilities, which can cause imbalances during model training or inference. For example, low-power sensors or devices with limited processing capacity may need help to participate in complex model training tasks, resulting in inefficiencies or delays. This heterogeneity makes it difficult to ensure that all devices can effectively collaborate in the model training process, particularly when deploying LMs that require substantial computational resources.

Furthermore, the varying capabilities of IoT devices complicate the task of synchronizing and optimizing model training across the entire network. High-performance devices may be underutilized if the workload is not properly distributed, while lower-resource devices could be overwhelmed by the demands of training. This mismatch in device capabilities necessitates strategies that can effectively manage and balance the computational load across devices to ensure that the system operates efficiently, even in large-scale IoT deployments with diverse hardware.

\subsection{Potential Solutions}
\subsubsection{Dynamic Resource Management}
Dynamic resource management offers a robust solution to minimize energy consumption while maintaining the IoT system's performance. This approach ensures efficient resource utilization across IoT devices and edge servers by dynamically adapting computational and communication workloads based on energy availability and task requirements.

One key strategy is dynamic GPU frequency scaling on edge servers. The system can balance computational performance and energy usage by adjusting GPU frequencies in real time based on workload demands. For example, lower frequencies can be used during idle periods, significantly reducing energy consumption without impacting latency-sensitive operations.

Another critical technique is energy-aware task scheduling, which allocates tasks to IoT devices based on their current energy levels and capabilities. High-energy tasks can be offloaded to devices with greater energy reserves or processed on the edge server, while low-energy devices handle simpler operations. Furthermore, compression of intermediate activations reduces the size of data transmitted between devices and servers, lowering communication energy costs. Together, these strategies optimize energy efficiency, making split knowledge distillation viable for resource-constrained IoT environments.

\subsubsection{Adaptive Model Partition}
To address the challenge of low training delay, a solution is to optimize the division of the model between IoT devices and the edge server. By dynamically adjusting the partition point (cut layer) of the model based on real-time network conditions, computational resources, and task requirements, this approach ensures minimal delay in both training and inference.

In this method, the device-side portion of the model handles the less computationally intensive layers, while the server-side portion processes the more complex parts of the model. The adaptive selection of the cut layer ensures that IoT devices with limited computational resources are not burdened with excessive processing, while the edge server can handle more complex computations, reducing latency. Moreover, this partitioning approach can adjust in response to changes in network bandwidth, device workload, and energy availability, enabling efficient use of resources and minimizing delays in training.

By optimizing the distribution of training tasks across the system, adaptive model partitioning enables faster model updates and real-time inference, meeting the stringent latency requirements of IoT applications without compromising the accuracy or efficiency of the distillation process.

\subsubsection{Clustered Collaborative Training}
To address the challenge of heterogeneous IoT devices, clustered collaborative training offers a practical solution by grouping IoT devices with similar resource capabilities into clusters. Within each cluster, devices can focus on model training tasks that match their computational power, allowing for more efficient use of resources. For example, high-performance devices in a cluster can handle more complex model layers, while lower-resource devices can focus on less demanding tasks, such as data preprocessing or training earlier model layers.

This approach ensures that all devices can effectively participate in training and reduces the communication overhead between devices and the edge server by limiting data exchange to relevant clusters. By optimizing task distribution and communication, clustered collaborative training enables efficient model training across heterogeneous IoT devices, ensuring that resource-constrained devices are not overburdened and that high-performance devices are fully utilized. This balanced approach helps maintain scalability and performance in IoT networks, allowing them to handle LMs effectively despite their diverse device ecosystem.

\section{Proposed Split Knowledge Distillation Framework} \label{sec: Proposed Framework}

\subsection{System Model}
Deploying LMs on IoT devices is challenging due to their limited computational and memory resources.  In addition, to protect data privacy, raw data generated on IoT devices cannot be directly transmitted to edge servers for model training. 
Hence, we propose a split knowledge distillation framework to  distill LMs into a smaller model, deployable versions on IoT devices while ensuring that raw data remains stored locally on the devices. In this framework, an edge server collaborates with IoT devices to jointly perform model training and distillation, and then these devices implement the distilled model for real-time inference. The edge server and IoT devices are illustrated as follows.

\begin{itemize}
	\item \textbf{Edge server:} The edge server is deployed on an access point, maintaining the complete teacher LM and student model to ensure comprehensive access to global model parameters.
    The access point is responsible for gathering network information, such as device computing capabilities and channel conditions, which supports the server in making decisions about the cut layer selection and energy management.
	
    \item \textbf{IoT devices:} IoT devices operate within the signal coverage area of the access point and serve as lightweight endpoints in the framework. These devices have significantly lower computing power and storage capacity. Hence, each device hosts only a partial segment of the teacher LM while deploying the complete student model. 
\end{itemize}

The teacher LM is partitioned into the embedding module and the remaining layers. The remaining layers are most computationally expensive and consist of subsequent transformer blocks and a task module. The edge server stores the entire teacher LM, while each IoT device only stores the embedding module. In the training process, each IoT device executes the embedding module of the teacher LM, while the edge server completes the training process of the remaining layers. By doing this, the raw data stored on edge devices does not need to be uploaded to the edge server, which prevents privacy leakage. The whole student model is deployed on both the edge server and devices due to its lightweight.

The objective of the proposed framework is to determine the optimal cut layer and the server's GPU computational frequency, so as to minimize the total energy consumption while meeting a low model training delay requirement.  Each IoT device trains the initial layers of the student model at the cut layer, while the edge server completes the remaining layer training. The detailed workflow of the proposed framework is presented in the following section.

\subsection{System Workflow}
\begin{figure*}[t]
	\renewcommand{\figurename}{Fig.}
	\centering
	\includegraphics[width=0.6\textwidth]{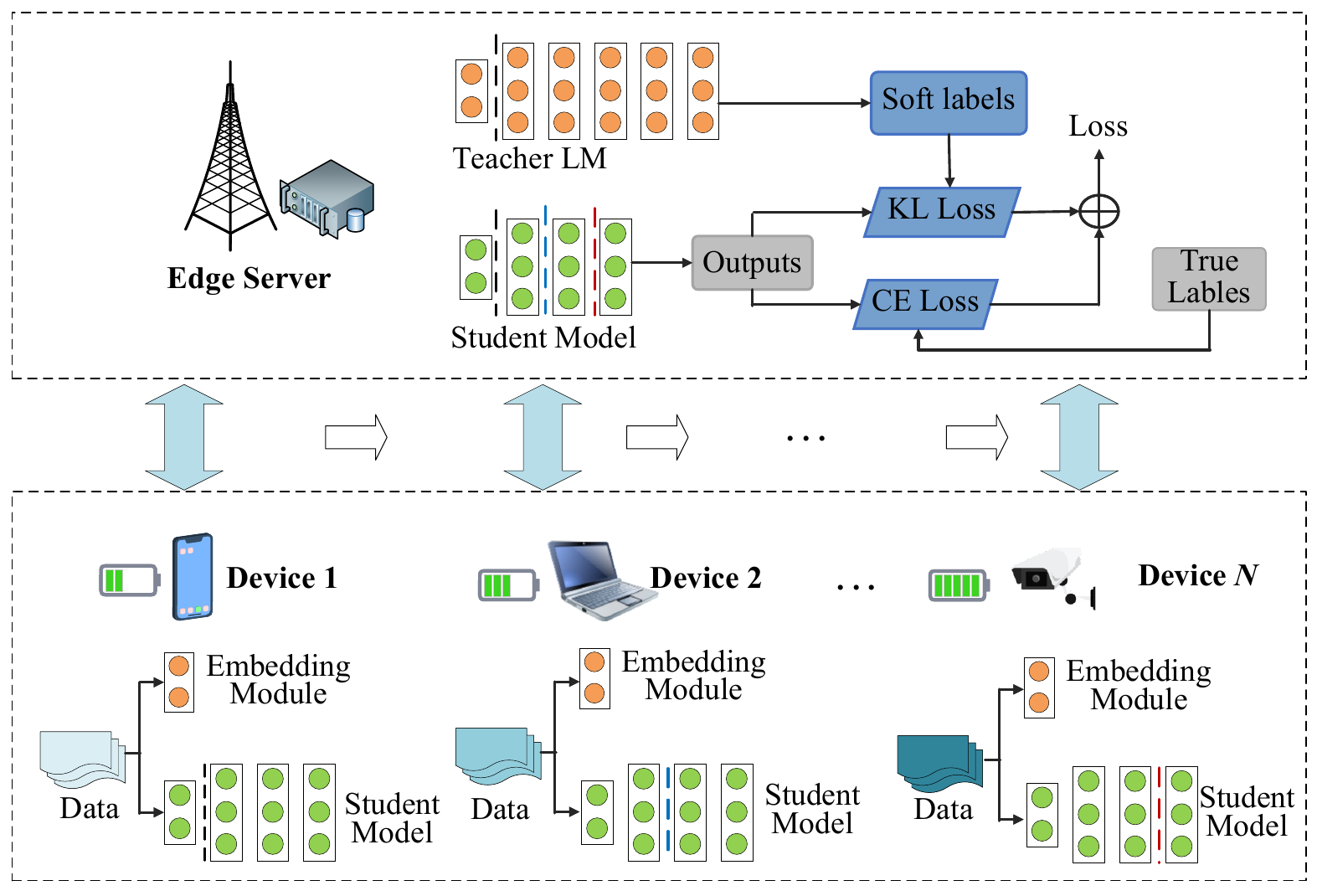}
	\caption{The training process of our proposed split knowledge distillation system.}
	\label{fig: The training process}
    \vspace{-0.4cm}
\end{figure*}

The edge server executes the model training with each IoT device in a line manner, as shown in Fig.~\ref{fig: The training process}. Several training rounds are performed between the server and these devices until a satisfactory performance of the student model is achieved. In a training round, the server may conduct several local epochs of model training with a device, where the number of local epochs depends on these factors such as the device's dataset size, computational capacity, and wireless channel conditions. After completing the local epochs, the device uploads the updated model parameters to the server.
The details of a training round are as follows.

The edge server dynamically selects an IoT device for collaborative training based on current wireless conditions and device resources. It determines the optimal cut layer between the device-side and server-side portions of the student model to balance energy, computation, and communication efficiency.
During training, the server transmits the index of the cut layer to the selected IoT device. The device processes its local data through the student model's initial layers and the teacher LM's embedding module, generating intermediate outputs, referred to as smashed data. These outputs are then transmitted to the edge server for further processing, ensuring raw data remains local to the device to preserve privacy.

The edge server completes the forward propagation using the remaining layers of the student model and the teacher LM. It computes a combined loss function consisting of KL divergence, which aligns the student model’s predictions with the teacher LM’s soft outputs, and cross-entropy loss, which ensures alignment with ground-truth labels. This joint loss function guides the training process to achieve both accuracy and compactness.

Once the loss gradients are computed, the edge server transmits them back to the IoT device, which performs local backpropagation to update the parameters of the device-side layers. This iterative process allows the device to refine its local model without requiring direct access to the full teacher LM or server-side data, significantly reducing communication overhead.
The server and devices repeat these training rounds collaboratively across all participating devices until the student model performs satisfactorily. This process ensures the distilled model is lightweight, accurate, and optimized for deployment in resource-constrained IoT environments.

\section{Case Study} \label{sec: Case Study}
\subsection{Considered Scenario}
To evaluate the performance of the proposed split knowledge distillation framework, we consider a realistic scenario where an edge server collaborates with 10 IoT-enabled vehicles to perform knowledge distillation. The task involves distilling a large LLaMA 3.2 8B model into a small LLaMA 3.2 1B model. The teacher LM comprises 64 transformer blocks, while the student model contains 12 transformer blocks.
This distilled model can support autonomous driving systems, offering functionalities such as text generation, knowledge graph construction, and real-time decision-making. The experimental setup assumes a single base station with an edge server, providing seamless connectivity to the 10 vehicles. These vehicles are equipped with heterogeneous IoT devices, showcasing varying computational capabilities. The computational resources of the edge server and the IoT devices are detailed in Table~\ref{Table: The Server and Devices Settings}. The edge server is a high-performance computer equipped with an Nvidia RTX 4090 GPU, capable of handling the intensive computational requirements of the large LLaMA model. On the other hand, the 10 vehicles are equipped with a diverse range of Jetson devices, reflecting the resource heterogeneity commonly found in real-world IoT environments.

As the vehicles operate within the base station's coverage area, they initiate model training tasks at the beginning of each trial. During the experiment, the vehicles follow a predefined trajectory at a constant speed of 30 km/h, simulating a typical urban driving scenario. 
The edge server and vehicles communicate over a 5G mmWave channel. The link bitrate is dynamically determined using the 5G New Radio (NR) CQI-to-MCS mapping table~\cite{3gpp2022phsical}, providing a realistic approximation of mobile cellular network conditions. To further emulate real-world dynamics, the framework is tested under three distinct channel quality conditions: \textit{Good}, \textit{Normal}, and \textit{Poor}, which correspond to noise spectral densities of -166 dBm/Hz, -163 dBm/Hz, and -160 dBm/Hz, respectively~\cite{wu2020accuracy}.  These conditions emulate the typical variations in signal quality encountered in real-world vehicular networks, allowing for a comprehensive assessment of the framework's robustness and adaptability to channel fluctuations.

\begin{table}[t]
	\scriptsize
    \renewcommand\arraystretch{1.5}
	\centering
	\caption{The Edge Server and IoT Devices Settings.}
	\label{Table: The Server and Devices Settings}
        \centering
	\begin{tabular}{cccc}
        \hline
		\hline
	   & \textbf{IoT Device}  & \textbf{Max GPU Frequency} & \textbf{Cores}\\
		\hline
        Edge Server &  \begin{tabular}[c]{@{}l@{}} A Computer with \\ Nvidia 4090 GPU \end{tabular} & 2.52 GHz & 16384 \\ \hline
	  Devices 1\&2 & Jetson AGX Orin & 1.3 GHz & 2048  \\
        Devices 3\&4 & Jetson AGX Orin & 1.0 GHz & 2048  \\
        Devices 5\&6 & Jetson Orin NX &  0.91 GHz & 1024  \\
        Devices 7\&8 & Jetson Orin NX & 0.76 GHz & 1024  \\
        Devices 9\&10 & Jetson AGX Xavier & 1.2 GHz & 512  \\
        \hline
	\end{tabular}
	\vspace{-0.3cm}
\end{table}

\subsection{Simulation Results}

We first evaluate the model training performance of the proposed approach in terms of training delay and energy consumption. We compared the proposed approach with two benchmark methods: (\textit{i}) Server-only, where devices train the first layer of the student model and the embedding module of the teacher LM locally, while the server completes the remaining training process. (\textit{ii}) Device-only, where devices train the first 11 layers of the student model and the embedding module of the teacher LM locally, while the server completes the remaining training process.

\begin{figure}[t]
	\renewcommand{\figurename}{Fig.}
	\centering
	\includegraphics[width=0.35\textwidth]{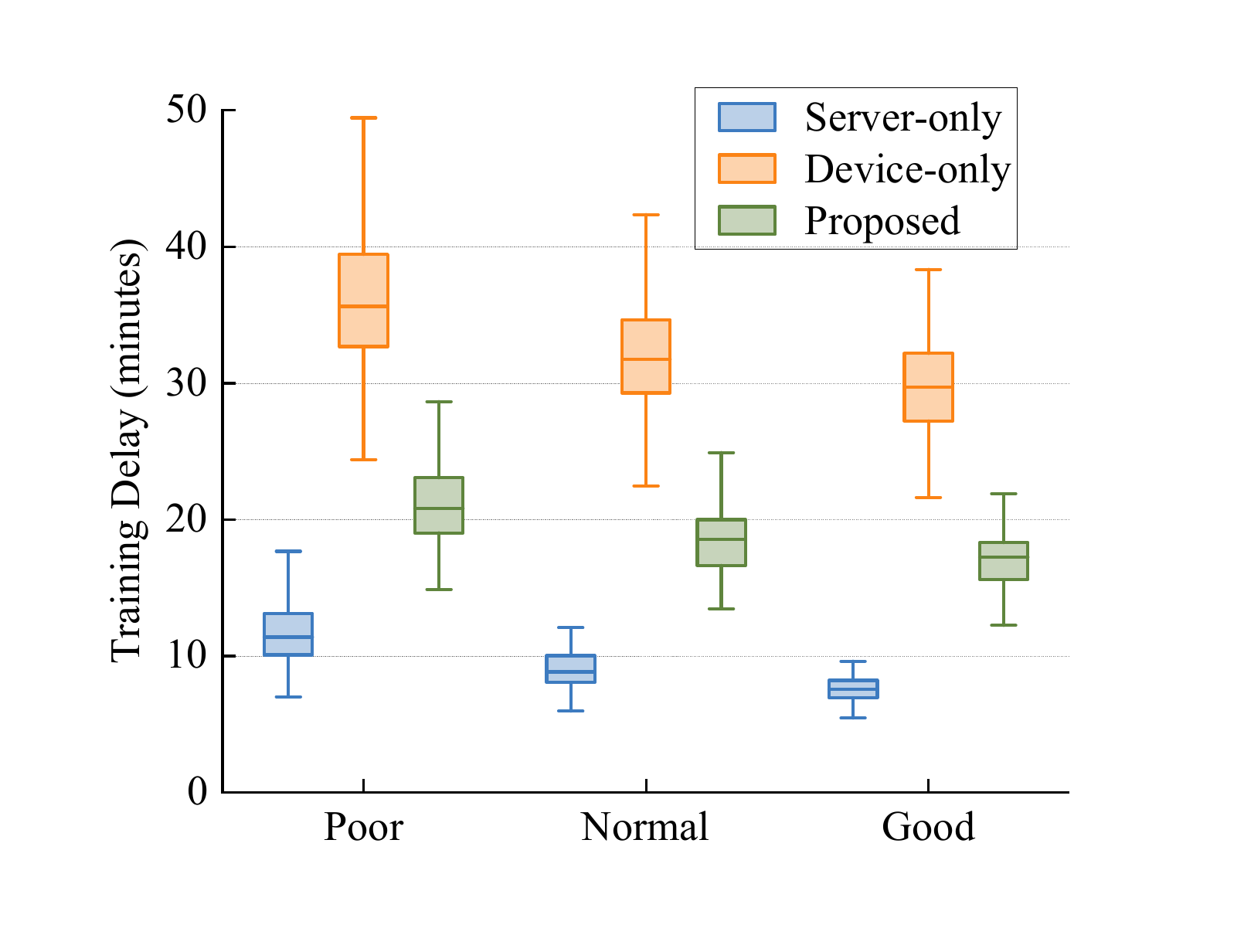}
	\caption{Comparison of training delay in different methods.}
	\label{fig: Training delay}
    \vspace{-0.4cm}
\end{figure}

In Fig.~\ref{fig: Training delay}, we show the training delay during a training round associated with each method under three channel conditions: \textit{Good}, \textit{Normal}, and \textit{Poor}. The results demonstrate that the proposed method achieves significantly lower training delay across all conditions compared to the server-only approaches. In particular, the training delay of the proposed method is reduced by at least 16\% compared to the server-only approach. This improvement is attributed to the efficient partitioning of computation workload in the proposed framework.

\begin{figure}[t]
	\renewcommand{\figurename}{Fig.}
	\centering
	\includegraphics[width=0.35\textwidth]{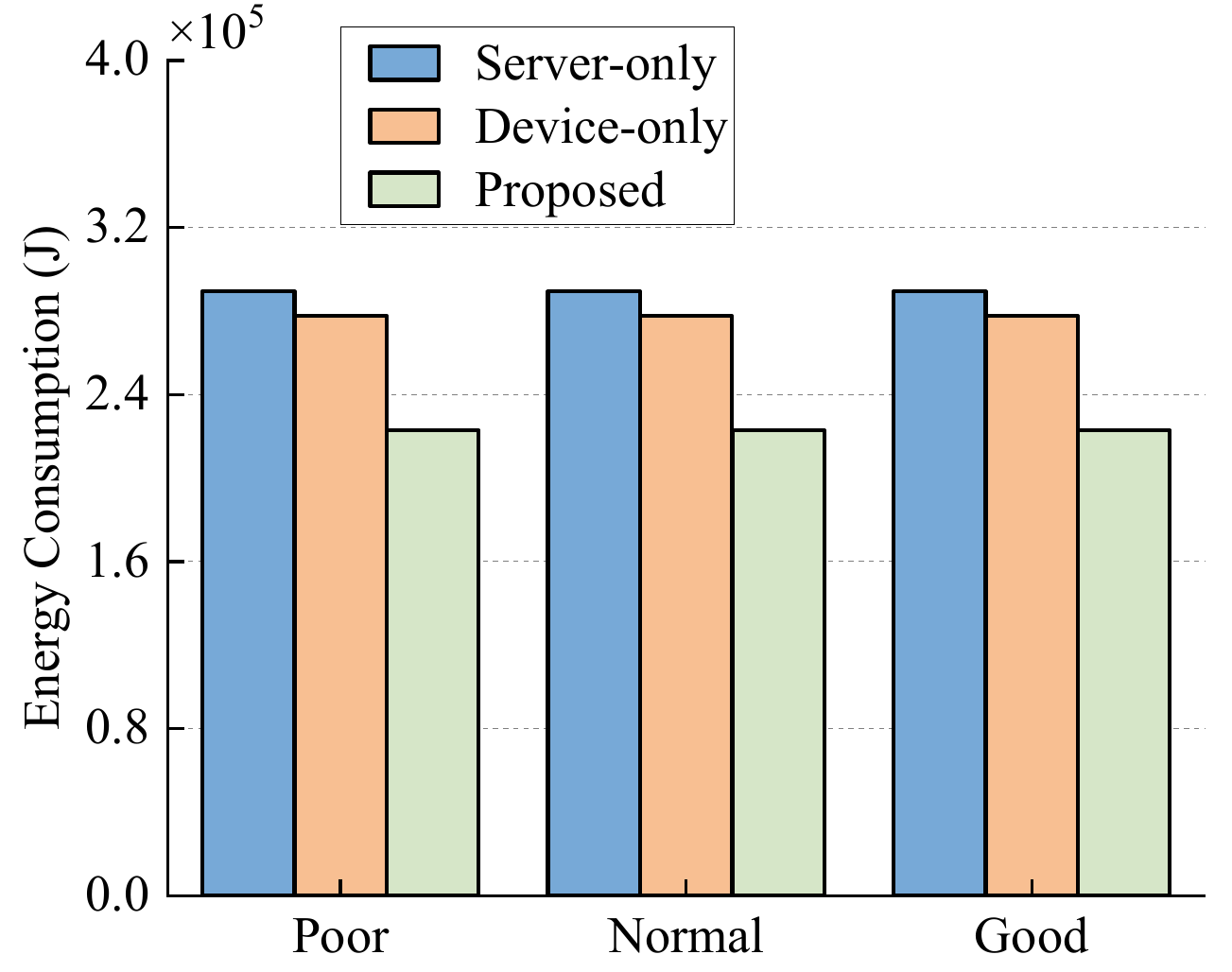}
	\caption{Comparison of average energy consumption in different methods.}
	\label{fig: Energy consumption}
    \vspace{-0.4cm}
\end{figure}

In addition to training delay, Fig.~\ref{fig: Energy consumption} illustrates the energy consumption during a training round under three channel conditions. The proposed method consistently exhibits the lowest energy consumption compared to the benchmark methods. The proposed approach consumes at least 22\% less energy than the server-only approach and 19\% less than the device-only approach. This energy efficiency is achieved through dynamic cut layer selection and effective resource management of the edge server, highlighting the suitability of the proposed method for energy-constrained IoT environments.

\section{Conclusion}  \label{sec: Conclusion}

We have explored the challenges and potential solutions for training LMs in IoT systems, emphasizing energy efficiency, low training latency, and heterogeneous device capabilities. To address these challenges, we have proposed a split knowledge distillation framework that integrates knowledge distillation and split learning. This framework efficiently distills LMs into smaller, deployable versions suitable for IoT devices while ensuring that raw data remains local to preserve privacy. Through a case study, we have validated the framework's feasibility and effectiveness, demonstrating its potential for enabling advanced AI applications in resource-constrained IoT environments.

\bibliographystyle{IEEEtran}
\bibliography{main}

\begin{thebibliography}{10}
\providecommand{\url}[1]{#1}
\csname url@samestyle\endcsname
\providecommand{\newblock}{\relax}
\providecommand{\bibinfo}[2]{#2}
\providecommand{\BIBentrySTDinterwordspacing}{\spaceskip=0pt\relax}
\providecommand{\BIBentryALTinterwordstretchfactor}{4}
\providecommand{\BIBentryALTinterwordspacing}{\spaceskip=\fontdimen2\font plus
\BIBentryALTinterwordstretchfactor\fontdimen3\font minus \fontdimen4\font\relax}
\providecommand{\BIBforeignlanguage}[2]{{%
\expandafter\ifx\csname l@#1\endcsname\relax
\typeout{** WARNING: IEEEtran.bst: No hyphenation pattern has been}%
\typeout{** loaded for the language `#1'. Using the pattern for}%
\typeout{** the default language instead.}%
\else
\language=\csname l@#1\endcsname
\fi
#2}}
\providecommand{\BIBdecl}{\relax}
\BIBdecl

\bibitem{cui2024llmind}
H.~Cui, Y.~Du, Q.~Yang, Y.~Shao, and S.~C. Liew, ``{LLMind: Orchestrating AI and IoT with LLM for complex task execution},'' \emph{IEEE Commun. Mag.}, 2024, \url{DOI:10.1109/MCOM.002.2400106}.

\bibitem{xiao2024efficient}
B.~Xiao, B.~Kantarci, J.~Kang, D.~Niyato, and M.~Guizani, ``Efficient prompting for {LLM}-based generative {Internet} of {Things},'' \emph{IEEE Internet Things J.}, 2024, \url{DOI:10.1109/JIOT.2024.3470210}.

\bibitem{na2024understanding}
S.~Na, G.~Jeong, B.~H. Ahn, J.~Young, T.~Krishna, and H.~Kim, ``Understanding performance implications of {LLM} inference on {CPUs},'' in \emph{Proc. IEEE IISWC}, 2024, pp. 169--180.

\bibitem{yang2024survey}
C.~Yang, Y.~Zhu, W.~Lu, Y.~Wang, Q.~Chen, C.~Gao, B.~Yan, and Y.~Chen, ``Survey on knowledge distillation for large language models: Methods, evaluation, and application,'' \emph{ACM Trans. Intell. Syst. Technol.}, 2024, \url{DOI:10.1145/3699518}.

\bibitem{wu2023split}
W.~Wu, M.~Li, K.~Qu, C.~Zhou, X.~Shen, W.~Zhuang, X.~Li, and W.~Shi, ``Split learning over wireless networks: Parallel design and resource management,'' \emph{IEEE J. Sel. Areas Commun.}, vol.~41, no.~4, pp. 1051--1066, 2023.

\bibitem{huang2023advancing}
Y.~Huang, J.~Xu, J.~Lai, Z.~Jiang, T.~Chen, Z.~Li, Y.~Yao, X.~Ma, L.~Yang, H.~Chen \emph{et~al.}, ``Advancing transformer architecture in long-context large language models: A comprehensive survey,'' \emph{arXiv:2311.12351}, 2023.

\bibitem{ahuja2021touchpose}
K.~Ahuja, P.~Streli, and C.~Holz, ``{TouchPose}: Hand pose prediction, depth estimation, and touch classification from capacitive images,'' in \emph{Proc. ACM UIST}, 2021, pp. 997--1009.

\bibitem{xu2023survey}
C.~Xu and J.~McAuley, ``A survey on model compression and acceleration for pretrained language models,'' in \emph{Proc. AAAI}, vol.~37, no.~9, 2023, pp. 10\,566--10\,575.

\bibitem{rao2024parameter}
J.~Rao, X.~Meng, L.~Ding, S.~Qi, X.~Liu, M.~Zhang, and D.~Tao, ``Parameter-efficient and student-friendly knowledge distillation,'' \emph{IEEE Trans. Multimedia}, vol.~26, pp. 4230--4241, 2024.

\bibitem{kang2017neurosurgeon}
Y.~Kang, J.~Hauswald, C.~Gao, A.~Rovinski, T.~Mudge, J.~Mars, and L.~Tang, ``Neurosurgeon: Collaborative intelligence between the cloud and mobile edge,'' \emph{ACM SIGARCH Computer Architecture News}, vol.~45, no.~1, pp. 615--629, 2017.

\bibitem{ayad2023efficient}
A.~Ayad, M.~Barhoush, M.~Frei, B.~V{\"o}lker, and A.~Schmeink, ``An efficient and private {ECG} classification system using split and semi-supervised learning,'' \emph{IEEE J. Biomed. Health Inform.}, vol.~27, no.~9, pp. 4261--4272, 2023.

\bibitem{padaria2023traffic}
A.~A. Padaria, A.~A. Mehta, N.~K. Jadav, S.~Tanwar, D.~Garg, A.~Singh, G.~Pau, and G.~Sharma, ``Traffic sign classification for autonomous vehicles using split and federated learning underlying {5G},'' \emph{IEEE Open J. Veh. Technol.}, vol.~4, pp. 877--892, 2023.

\bibitem{ghosh2024split}
B.~Ghosh, Y.~Wang, H.~Fu, Q.~Wei, Y.~Liu, and R.~S.~M. Goh, ``Split learning of multi-modal medical image classification,'' in \emph{Proc. IEEE CAI}, 2024, pp. 1326--1331.

\bibitem{3gpp2022phsical}
{3GPP}, ``{NR; Physical layer procedures for data},'' document TS 38.214 V17.3.0, Sep. 2022.

\bibitem{wu2020accuracy}
W.~Wu, P.~Yang, W.~Zhang, C.~Zhou, and X.~Shen, ``{Accuracy-guaranteed collaborative DNN inference in industrial IoT via deep reinforcement learning},'' \emph{IEEE Trans. Ind. Informat.}, vol.~17, no.~7, pp. 4988--4998, 2020.

\end{thebibliography}

\end{document}